\pdfoutput=1
\documentclass{article}
\usepackage[final]{neurips_2021}
\usepackage[utf8]{inputenc} %
\usepackage[T1]{fontenc}    %
\usepackage{hyperref}       %
\usepackage{url}            %
\usepackage{booktabs}       %
\usepackage{amsfonts}       %
\usepackage{nicefrac}       %
\usepackage{microtype}      %
\usepackage{graphicx}
\usepackage{multirow}
\usepackage{siunitx}
\usepackage{subfig}

\title{Lesan -- Machine Translation for Low Resource Languages}

\author{%
  Asmelash Teka Hadgu \\
  \texttt{asme@lesan.ai} \\
  \And
  Abel Aregawi \\
  \texttt{abel@lesan.ai} \\
  \And
  Adam Beaudoin\\
  \texttt{adam@lesan.ai} \\
}

\begin{document}

\maketitle

\begin{abstract}

	Millions of people around the world can not access content on the Web because most of the content is not readily available in their language. Machine translation (MT) systems have the potential to change this for many languages. Current MT systems provide very accurate results for high resource language pairs, e.g., German and English. However, for many low resource languages, MT is still under active research. The key challenge is lack of datasets to build these systems. We present Lesan~\footnote{\url{https://lesan.ai/}}, an MT system for low resource languages. Our pipeline solves the key bottleneck to low resource MT by leveraging online and offline sources, a custom Optical Character Recognition (OCR) system for Ethiopic and an automatic alignment module. The final step in the pipeline is a sequence to sequence model that takes parallel corpus as input and gives us a translation model. Lesan's translation model is based on the Transformer architecture. After constructing a base model, back translation is used to leverage monolingual corpora. Currently Lesan supports translation to and from Tigrinya, Amharic and English. We perform extensive human evaluation and show that Lesan outperforms state-of-the-art systems such as Google Translate and Microsoft Translator across all six pairs. Lesan is freely available and has served more than 10 million translations so far. At the moment, there are only 217 Tigrinya and 15,009 Amharic Wikipedia articles. We believe that Lesan will contribute towards democratizing access to the Web through MT for millions of people.

\end{abstract}

\section{Machine Translation Pipeline}

\begin{figure}[!ht]
  \centering
  \includegraphics[width=\linewidth]{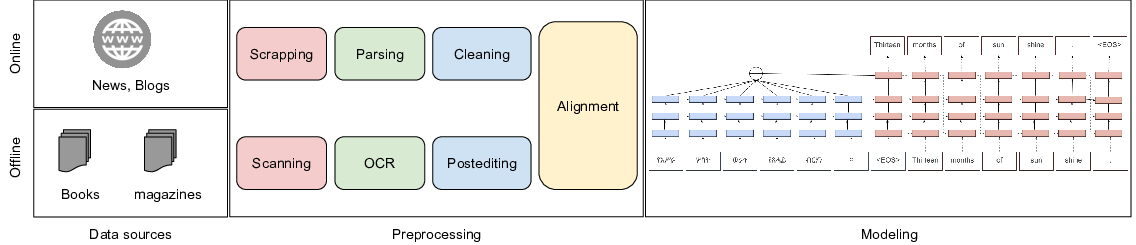}
  \caption{Lesan machine translation system pipeline}
  \label{fig:pipeline}
\end{figure}

Our machine translation pipeline shown in Figure~\ref{fig:pipeline} has the
following three key components: OCR, automatic alignment and MT
modeling that leverage current advances in deep learning.

\paragraph{Data sources} Machine translation (MT) systems today provide very
accurate results for high resource language pairs~\cite{barrault2019findings}.
However, these systems rely on large-scale parallel corpus to work. For pairs
such as German and English, these parallel sentences can be easily obtained
from the Web. However, for the majority of languages there is not enough data
online to build these datasets. We leverage offline and online sources to
mitigate this bottleneck.

Offline sources are gathered through partnerships with media companies,
libraries and other archives that have books, magazines and other linguistic
materials. These are typically translated books or magazines that have daily
news of source and target language in different file formats and hard copies.
Online sources include news sites, blogs, religious and legal sites that
feature stories in both the source and target language.

\paragraph{Data preprocessing}

After gathering large-scale unstructured data from online and offline sources,
the next step is a preprocessing step. The goal of this module is to generate
clean text (corpus) in machine readable format.
Scanned books and magazines pass through an OCR
system to convert them into machine readable format. We built a custom OCR
system for Ethiopic, script used for Tigrinya, Amharic and several other
languages.
The output of our OCR system is passed through post-editing to correct errors
in characters and words.
Data crawled from the Web is parsed to extract text by removing HTML-tags and
gathering metadata such as author, date that could be helpful in downstream
steps.
Text coming from different sources such as text parsed from web pages and text
coming from the post-editing module passed through several NLP modules such as
deduplication, language detection and sentence splitting.

\paragraph{Automatic alignment}

All source and target sentences that have been cleaned are then passed through
an automatic alignment system to construct parallel corpus. 
A candidate generation step, produces candidate pairs, then each pair of source
and target sentence is fed to our automatic matching system that determines
whether an input pair is a translation or not. This system is similar
to~\cite{junczys2018dual} where we compute cross-entropy scores according to two
inverse translation models, and take the weighted average aggregated over a set
of candidate pairs.

\paragraph{Sequence-to-sequence modeling with back translation}

The final step in the pipeline is a sequence to sequence model that takes the
parallel corpus as input and gives us a translation model. Lesan's translation
model is based on the Transformer architecture~\cite{vaswani2017attention}.
After constructing a base model, back
translation~\cite{edunov2018understanding} is used to leverage monolingual
corpora.

Our machine translation pipeline solves the key bottleneck to low resource MT
by leveraging online and offline sources, a custom OCR system for Ethiopic and
an automatic alignment module.

\section{Evaluation} 

In this section we will describe human evaluation of Lesan compared to three
state-of-the-art commercial systems: Google
Translate~\footnote{\url{https://translate.google.com/}}, Microsoft
Translator~\footnote{\url{https://www.bing.com/translator}} and Yandex
Translate~\footnote{\url{https://translate.yandex.com/}}.
Google Translate uses neural MT~\citep{wu2016google}. Yandex Translate uses a
hybrid system of neural MT and statistical
MT~\footnote{\url{https://tech.yandex.com/translate/doc/dg/concepts/how-works-machine-translation-docpage/}}.
All these services provide APIs to access their system. Lesan's MT models are
implemented using the OpenNMT~\cite{klein2017opennmt} toolkit.

\paragraph{Human Evaluation}

Experts were selected to evaluate the translation outputs of the systems. We
assign the task of selecting what news sources to translate to the expert
evaluators in order to avoid bias. The main requirements were: identify 20 news
stories (1 - 2 paragraphs); select stories across different genres: politics,
social, economy, entertainment, health, sport, technology etc.
The output of translation systems next to the sources are given to expert
evaluators to assign a score. The outputs are shuffled in such a way that one
cannot tell which output is from which system.
We chose a Likert
scale~\footnote{\url{https://en.wikipedia.org/wiki/Likert\_scale}} of 5
corresponding to a range from completely wrong (0) to accurate and fluent (4).
These error categories are adapted from the severity levels for error
categories
~\footnote{\url{https://www.taus.net/qt21-project\#harmonized-error-typology}}
used in evaluation of translation quality developed by TAUS. The complete
description of the scoring scheme is shown in Table~\ref{tab:eval-guide}. 
We have released the evaluation
datasets~\footnote{\url{https://zenodo.org/record/5060303}} to
foster research and progress on evaluating MT systems for low resource
languages.

\begin{center}
	\begin{table*}[!ht]
\small
\caption{Human evaluation guideline to evaluate performance of MT systems.}
\label{tab:eval-guide}

\begin{tabular}{ c | c | p{10cm} }
Scale & Value & Description \\
\hline
Wrong translation & 0 & This is for a completely wrong translation. The translation output does not make sense given the source text. \\
Major problem & 1 & There is a serious problem in the translation with some parts of the source missing or mistranslated and it would be hard to match translation output with source text without major modifications. \\
Minor problem & 2 & The translation has minor problems given the source text but requires some minor changes, e.g, changing a word or two to make it fully describe the source text. \\
Good translation & 3 & The translation describes the source text; however, there may be some problems with style such as punctuation, word order or appropriate wording. \\
Accurate and fluent & 4 & Great job! The output is a correct translation of the source text. It’s both accurate and fluent.
\end{tabular}
\end{table*}
\end{center}

We report the normalized mean and standard deviation of the scores. The results
are given in Table~\ref{tab:eval-am} for Amharic to and from English and in
Table~\ref{tab:eval-ti} for Tigrinya to and from Amharic and English. Across
all directions Lesan outperforms these state-of-the-art systems.

\begin{table*}[!ht]
\parbox{.45\linewidth}{
\caption{Human evaluation comparing Lesan against three commercial MT systems for Amharic to and from English.}
\label{tab:eval-am}
\begin{center}
\begin{tabular}{r|r|c|c}
Direction & System & Sentence & Story \\
\hline
	\multirow{4}{*}{\small{Am $\rightarrow$ En}} & Yandex    & 0.23 \tiny{$\pm$ 0.30}  & 0.19 \tiny{$\pm$ 0.25} \\
                                     & Microsoft & 2.13 \tiny{$\pm$ 0.51}  & 2.06 \tiny{$\pm$ 0.50} \\
                                     & Google    & 2.58 \tiny{$\pm$ 0.54}  & 2.54 \tiny{$\pm$ 0.48} \\
		                     & Lesan     & \textbf{2.68} \tiny{$\pm$ 0.41}  & \textbf{2.71} \tiny{$\pm$ 0.55} \\
\hline
	\multirow{4}{*}{\small{En $\rightarrow$ Am}} & Yandex    & 0.28 \tiny{$\pm$ 0.34}  & 0.20 \tiny{$\pm$ 0.29} \\
                                     & Microsoft & 2.57 \tiny{$\pm$ 0.43}  & 2.54 \tiny{$\pm$ 0.44} \\
                                     & Google    & 2.98 \tiny{$\pm$ 0.30}  & 2.88 \tiny{$\pm$ 0.33} \\
                                     & Lesan     & \textbf{3.25} \tiny{$\pm$ 0.38}  & \textbf{3.17} \tiny{$\pm$ 0.42} \\

\end{tabular}
\end{center}
}
\hfill
\parbox{.45\linewidth}{
\caption{Human evaluation comparing Lesan and Microsoft for Tigrinya to and from Amharic and English.}
\label{tab:eval-ti}
\begin{center}
\begin{tabular}{r|r|c|c}
Direction & System & Sentence & Story \\
\hline
	\multirow{2}{*}{\small{Am $\rightarrow$ Ti}} & Microsoft  & 1.92 \tiny{$\pm$ 0.43}  & 1.85 \tiny{$\pm$ 0.54} \\
		                     & Lesan      & \textbf{1.94} \tiny{$\pm$ 0.47}  & \textbf{1.86} \tiny{$\pm$ 0.53} \\
\hline
	\multirow{2}{*}{\small{Ti $\rightarrow$ Am}} & Microsoft  & 1.60 \tiny{$\pm$ 0.44}  & 1.44 \tiny{$\pm$ 0.57} \\
                                     & Lesan      & \textbf{1.94} \tiny{$\pm$ 0.50}  & \textbf{1.77} \tiny{$\pm$ 0.50} \\
\hline
	\multirow{2}{*}{\small{En $\rightarrow$ Ti}} & Microsoft  & 2.32 \tiny{$\pm$ 0.6}  & 2.17 \tiny{$\pm$ 0.63} \\
		                     & Lesan      & \textbf{2.33} \tiny{$\pm$ 0.63}  & \textbf{2.19} \tiny{$\pm$ 0.58} \\
\hline
	\multirow{2}{*}{\small{Ti $\rightarrow$ En}} & Microsoft  & 2.01 \tiny{$\pm$ 0.63}  & 1.89 \tiny{$\pm$ 0.67} \\
                                     & Lesan      & \textbf{2.78} \tiny{$\pm$ 0.31}  & \textbf{2.63} \tiny{$\pm$ 0.39} \\

\end{tabular}
\end{center}
}
\end{table*}

\section{Broader Impact}

There are several applications of machine translation systems for broader impact. Let's take the case of Wikipedia. Wikipedia currently has a total of more than six billion articles and over 17 billion words in its English edition. Unfortunately, millions of people cannot access this because it's not available in their language. For instance, at the moment there are only 217 articles on the Tigrinya Wikipedia and 15,009 articles on the Amharic Wikipedia~\footnote{\url{https://en.wikipedia.org/wiki/List_of_Wikipedias}}. In future work, we would like to leverage Lesan's MT system to empower human translators towards our mission of opening up the Web's content to millions of people in their language.

\begin{ack}
We would like to thank Sergey Edunov from Facebook AI Research for valuable feedback on our machine translation pipeline.
\end{ack}

\bibliographystyle{unsrtnat}
\bibliography{main}

\appendix

\begin{figure}[!ht]
  \centering
  \includegraphics[width=\linewidth]{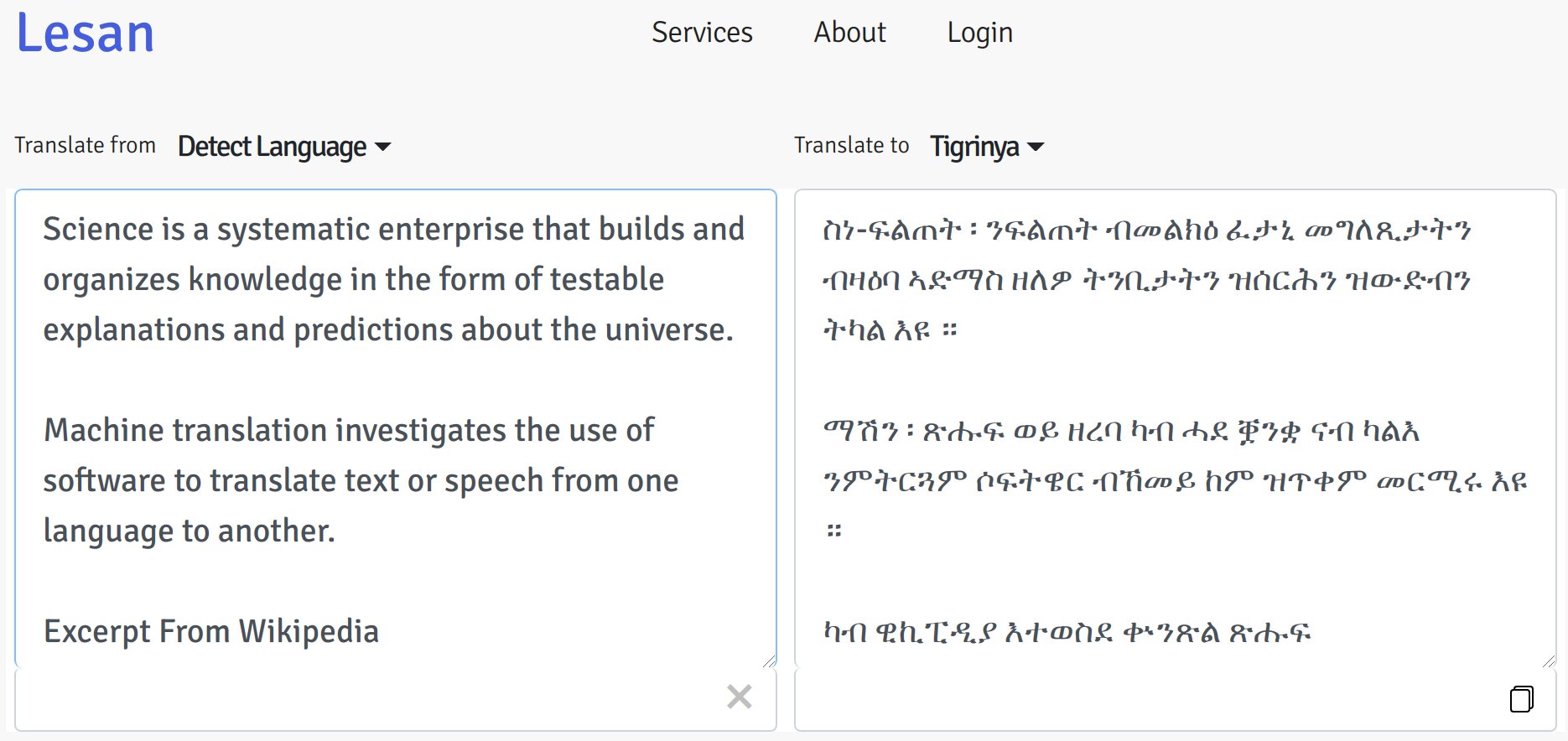}
  \caption{Web interface of Lesan - on the left is an excerpt from English Wikipedia and on the right the translation output to Tigrinya.}
  \label{fig:lesan-enti}
\end{figure}

\end{document}